\documentclass[runningheads]{llncs}
\usepackage{dcolumn}
\usepackage{graphics} 
\usepackage{mathptmx} 
 \usepackage{graphicx}
\usepackage{color, colortbl}
\usepackage{xcolor}
\usepackage{graphicx}
\usepackage{longtable}
\usepackage{verbatim}
 \usepackage{url}
\usepackage{tikz}
\usepackage{tikz-qtree}
\usetikzlibrary{shapes,shapes.geometric,arrows,fit,calc,positioning}
\usepackage{amsmath}
\usepackage{listings}
\definecolor{lightGray}{gray}{0.9}

\begin{document}
\title{Solving probability puzzles with logic toolkit}
%
%
\author{Adrian Groza\inst{1}\orcidID{0000-0003-0143-5631}}
\authorrunning{A. Groza}
%

\institute{Technical University of Cluj-Napoca, \and
Cluj-Napoca, Romania
\email{Adrian.Groza@cs.utcluj.ro}\\
\url{http://users.utcluj.ro/~agroza} 
}
\maketitle              
%

%
%
%

Solving logic puzzles by modelling them in first order logic (FOL) is old and recurrent hat in computer science~\cite{hazzan2020guide}. 
For instance, Prover9 theorem prover and Mace4 model finder~\cite{mccune2003mace4} have been applied to logical puzzles~\cite{groza2021modelling}.
Yet, the question is weather the same logic toolkit can be applied to the task of solving probabilistic puzzles. 
Here is an example for you: \textit{($Puzzle_1$) Let two decks of cards. Turn over the top card on each deck. What is the
probability that at least one of those cards is the queen of hearts?}

The proposed approach is to formalise the probabilistic puzzle in equational FOL. 
Two formalisations are needed: one theory for all models of the given puzzle, and a second theory for the favorable models. 
Then Mace4 - that computes all the interpretation models of a FOL theory -  is called twice. 
First, it is asked to compute all the possible models $\mathcal{M}_p$. 
Second, the additional constraint is added, and Mace4 computes only favorabile models $\mathcal{M}_f$.
Finally, the definition of probability is applied: the number of favorable models is divided by the number of possible models.  

$Puzzle_1$ is formalised in Listing~\ref{lst:two_decks_all}. 
The domain size is set to the first 52 natural numbers: i.e., all the functions from the theory are mapped against integer values from the interval [0..51]. 
With \texttt{max\_models=-1}, we are asking for all interpretations of the FOL theory.
Let the functions $deck_1$ and $deck_2$ take two values $a$ and $b$ from the domain. 
Mace4 computes $\mathcal{M}_p$=2,704 models, given by $52\times 52$.
Let queen of hearts be the value 7 (or anyother value) from the deck [0..51]. 
By adding the constraint in line 2 from Listing~\ref{lst:two_decks_fav}, Mace4 returns $\mathcal{M}_f$=103 models. 
Hence, the probability that at least one of those cards is the queen of hearts is $\frac{103}{2,704}$. 
This modelling-based approach avoids two common logical faults: 
(i) not considering the case of extracting both queens of heart with the possible wrong answer $\frac{1}{52}+ \frac{1}{52}=\frac{1}{26}$;
(ii) considering ordered pairs, where pairs like (Qh, Kh) are different from (Kh, Qh), with the wrong answer $\frac{52+52}{52\times52} = \frac{104}{2,704}$. 
\begin{minipage}{\linewidth}
\begin{lstlisting}[label=lst:two_decks_all,caption= Possible models when extracting two random cards]
assign(domain_size,52).
assign(max_models, -1).
set(arithmetic).
formulas(two_decks).
  deck1 = a.  
  deck2 = b. 
end_of_list.


\end{lstlisting}
\end{minipage}
\begin{minipage}{\linewidth}
\begin{lstlisting}[label=lst:two_decks_fav,caption= Favorable models for obtaining at leas tone queen of hearts]
formulas(Qh).
  a = 7 | b = 7.
end_of_list.


\end{lstlisting}
\end{minipage}

The above method is easily applied to dice puzzles. 
Let \textit{$Puzzle_2$: What is the probability of rolling three six-sided dice and getting a value greater
than 7?}. 
Let $Dice_1$, $Dice_2$, and $Dice_3$ the three dice (Listing~\ref{lst:3dice_all}). 
We set a domain size of 7 and we limit the possible values to 1 and 6 with $Dice_1\neq 0$, $Dice2 \neq 0$ and $Dice_3 \neq 0$. 
For three variables in [1..6], Mace4 returns $6\times 6 \times 6 = 216$ models, as expected. 
To compute the favourable cases, we add the constraint $Dice_1 + Dice_2 + Dice_3 > 7$ (Listing~\ref{lst:3dice_7}). 
Mace4 returns 181 models. 
Hence the probability is $\frac{181}{216}$.\\
\begin{minipage}{\linewidth}
\begin{lstlisting}[label=lst:3dice_all,caption= Possible models when throwing three dice]
assign(domain_size,7).
assign(max_models,-1).
set(arithmetic).
formulas(three_dice).
 Dice1 != 0.  Dice2 != 0.  Dice3 != 0.
end_of_list. 	
\end{lstlisting}
\end{minipage}
\begin{minipage}{\linewidth}
\begin{lstlisting}[label=lst:3dice_7,caption= Favorable models for a value greater than 7]
formulas(three_dice_favorable).
  Dice1 + Dice2 + Dice3 > 7. 
end_of_list. 	
\end{lstlisting}
\end{minipage}

Many logic and probabilistic puzzles are storified. Let $Puzzle_3$: 
\textit{A swindler once approached an honest man with a die. 
He handed him the die and told him about the bet. The die had six sides. If the man rolled 1, he wins, and gets back twice the amount of his bet. If not, the swindler would keep the bet.
"But...my chances are only one out of six," retorted the man.
"True," grinned the swindler, "But I'll give you three tries to get a one."
The man considered. Three tries, with each try having a 1/6 chance of winning. So his chances of winning are 1/2. Why not give it a try?
Is the bet really fair? If not, what are the chances of the man winning?}

For $Puzzle_3$, we reused the formalisation of possible models in Listing~\ref{lst:3dice_all}.
The favorable models require that at least one of the throwings is one 
(Listing~\ref{lst:swindler_fav}). 
For this, Mace4 computes 91 models. Hence the probability is 91/216, which is not a fair bet. \\
\begin{minipage}{\linewidth}
\begin{lstlisting}[label=lst:swindler_fav,caption= Is it a fair bet?]
formulas(swindler_fav).
 Dice1 = 1 | Dice2 = 1 | Dice3 = 1. 
end_of_list. 	

 
\end{lstlisting}
\end{minipage}

In Mace4, FOL contains predicates, functions (including arithmetic), quantifiers, which provides enough expressivity to formalise different puzzles, 
including those with given probabilities.
Let $Puzzle_4$: 
\textit{Your task is to select socks from a dark drawer. 
There are six socks, a mixture of black and white. 
If two socks a repicked, the chances that a white pair is drawn are 2/3. 
What are the chances that a black pair is drawn?}

Since there are six socks, we set the domain size to 6. 
Let the function $s(x)= 1$ if the sock $x$ is white and $s(x)= 0$ of the sock $x$ is black.
Let $W$ the number of white socks. 
Thus $\sum_{i=0}^5 s(i) = W$ (line 8 in Listing~\ref{lst:prob:socks_all}).  
The probability of drawing two white socks is: $W/6 *  (W-1)/5 = 2/3$. 
This is equivalent to $3 * W * (W-1) = 6 * 5 * 2$ (line 9 in Listing~\ref{lst:prob:socks_all}). 
Mace4 computes 6 possible models (right part of Listing~\ref{lst:prob:socks_all}).
We are interested in favorable models in which there are two socks $x\neq y$ such that $s(x) = Black$ and $sock(y) = Black$ (line 2 in Listing~\ref{lst:prob:socks_fav}).
For this constraint, Mace4 fails to compute any model. 
Hence, the probability of extracting two black socks is zero.
That is because all six possible models contain only one black sock.

\begin{minipage}{0.72\linewidth}
\begin{lstlisting}[label=lst:prob:socks_all, caption= Possible models for socks]
assign(domain_size,6).
assign(max_models,-1).
set(arithmetic).
formulas(socks_all).
 White = 1. Black = 0.
 s(x) = White | s(x) = Black. 
 s(0) + s(1) + s(2) + s(3) + s(4) + s(5) = W.  
 3 * W * (W + -1) =  6 * 5 * 2.
end_of_list.
\end{lstlisting}
\end{minipage}
\begin{minipage}{0.28\linewidth}
\centering
\begin{tabular}{l|l}
 Model &  $s(x)$ \\
  & 0 1 2 3 4 5\\ \hline
1 & 1 1 1 1 1 0\\
2 & 1 1 1 1 0 1\\
3 & 1 1 1 0 1 1\\
4 & 1 1 0 1 1 1\\
5 & 1 0 1 1 1 1\\
6 & 0 1 1 1 1 1\\
\end{tabular}
\end{minipage}
\begin{minipage}{\linewidth}
\begin{lstlisting}[label=lst:prob:socks_fav, caption= Models of two black socks]
formulas(socks_fav).
 exists x exists y (x != y & s(x) = Black & s(y) = Black).
end_of_list.

\end{lstlisting}
\end{minipage}

There is more than one formalisation for each given puzzle. 
Let the friendly $Puzzle_5$: \textit{Ross, Chandler, Joey, Monica and Rachel are celebrating at their favorite restaurant. 
While they are sitting down at a round table for five, Ross notes: ”We are siting down around the table in age order!
Which is the probability of that?}

To model the puzzle in FOL, we introduce the function $age(x)$ which should 
have distinct values for all variables in the domain: i.e., all five friends have different ages (line 6 in Listing~\ref{lst:prob:roundtable2_all}). 
To simulate the round table, let the successor function $s(x)$, where successor of the last place is the first place: $s(4) = 0$ (line 7 in Listing~\ref{lst:prob:roundtable2_all}).  
Mace4 computes 120 possible models. 
For favorable models, between five friends, there should be four order relations for ascending ages (lines 2-5 in Listing~\ref{lst:prob:roundtable2_fav}) and 
four relations for descending ages (lines 6-9).
The order relation - either ascending or descending - should hold for all friends (line 11 in Listing~\ref{lst:prob:roundtable2_fav}). 
Mace4 computes 10 favorable models. \\
\begin{minipage}{\linewidth}
\begin{lstlisting}[label=lst:prob:roundtable2_all, caption= Possible models for round table in FOL]
assign(domain_size,5).
assign(max_models,-1).
set(arithmetic).
formulas(round_table2_all).
 age(x) = age(y) -> x = y. 
 s(0) = 1. s(1) = 2. s(2) = 3. s(3) = 4. s(4) = 0. 
end_of_list.
\end{lstlisting}
\end{minipage}

We can improve the formalisation in FOL with the solution in Listing~\ref{lst:prob:roundtable2_fav}. 
Two auxiliary predicates $a(x)$ and $b(x)$ are true when $x$ and its successor $s(x)$ are in ascending order or descending order.
Also, the disjunction $order(0) \vee order(1) \vee order(2) \vee order(3) \vee order(4)$ is replaced by $\exists x\ order(x)$. \\ 
\begin{minipage}{\linewidth}
\begin{lstlisting}[label=lst:prob:roundtable2_fav, caption= Favorable models for round table in FOL]
formulas(round_table2_fav).
 ascending(y)  <-> age(y)             < age(s(y)) & 
                   age(s(y))          < age(s(s(y))) & 
                   age(s(s(y)))       < age(s(s(s(y)))) & 
                   age(s(s(s(s(y))))) < age(s(s(s(s(s(y)))))).
 descending(y) <-> age(y)             > age(s(y)) & 
                   age(s(y))          > age(s(s(y))) & 
                   age(s(s(y)))       > age(s(s(s(y)))) & 
                   age(s(s(s(s(y))))) > age(s(s(s(s(s(y)))))).
 order(y) <-> ascending(y) | descending(y).
 order(0) | order(1) | order(2) | order(3) | order(4). 
end_of_list.
\end{lstlisting}
\end{minipage}
\begin{minipage}{\linewidth}
\begin{lstlisting}[label=lst:prob:roundtable3_fav, caption= Improving FOL representation of the round table puzzle]
formulas(round_table3_fav).
 a(x)          <-> age(x) < age(s(x)).
 b(x)          <-> age(x) > age(s(x)).
 ascending(y)  <-> a(y) & a(s(y)) & a(s(s(y))) & a(s(s(s(y)))).
 descending(y) <-> b(y) & b(s(y)) & b(s(s(y))) & b(s(s(s(y)))).
 order(y)      <-> ascending(y) | descending(y).
 exists x order(x).
end_of_list.
\end{lstlisting}
\end{minipage}

The proposed approach equips students from the logic tribe to find the correct solution for puzzles from the probabilitistic tribe, 
by using their favourite instruments: modelling and formalisation. 
With this method, the student is trained to approach the task in terms of possible models and favorable models. 
Since the definition of probability never fails, pitfalls usually leading to so many distinct results and apparently correct to the learners are avoided. 
One challange is how far can one go with this approach. 
The optimistic hypothesis is that all probabilistic puzzles can be reduced to the definition of probability. 
Hence, a tool like Mace4 can handle probabilistic puzzles.    
A better question would be: what type of probabilistic puzzles are difficult to be modelled in FOL?

I have exemplified here five probabilistic puzzles and how they can be solved by translating them in FOL and then find the corresponding interpretation models. 
Mace4 was the tool of choice here.
Ongoing work is investigating the limits of this method on various collections of 
probabilistic puzzles from or from popular brain teaser sites like \url{https://www.brainzilla.com/} or \url{https://www.mathsisfun.com/}, or 
collections including 11 probabilistic puzzles in~\cite{velleman2020bicycle}, 56 puzzles in~\cite{mosteller1987fifty},~\cite{pearl1988probabilistic} or the larger collection of dice problems of Matthew M. Conroy~\cite{conroycollection}.

 \bibliographystyle{splncs04}
 \bibliography{main}
\end{document}